\pdfoutput=1

\documentclass[11pt]{article}

\usepackage{EMNLP2023}

\usepackage{times}
\usepackage{latexsym}

\usepackage[T1]{fontenc}

\usepackage[utf8]{inputenc}

\usepackage{microtype}

\usepackage{inconsolata}

\usepackage{ascii}
\usepackage{hyperref}
\usepackage{amsmath}
\usepackage{amsfonts}
\usepackage{algorithm}
\usepackage[noend]{algpseudocode}
\usepackage{bm}
\usepackage{bbm}
\usepackage{booktabs}
\usepackage{balance}
\usepackage{color,colortbl}
\usepackage{CJKutf8}
\usepackage{graphicx} 
\usepackage{graphics}
\usepackage{multirow}
\usepackage{makecell}
\usepackage{paralist}
\usepackage{subfig}
\usepackage{textcomp}
\usepackage{threeparttable}
\usepackage[normalem]{ulem}
\usepackage{arydshln}
\usepackage{enumitem}

\usepackage{cleveref}
\crefformat{section}{\S#2#1#3} 

\definecolor{jred}{RGB}{225, 11, 11}
\definecolor{jblue}{RGB}{41, 52, 190}
\definecolor{jgreen}{RGB}{18, 141, 21}
\definecolor{jorange}{RGB}{255, 127, 80}
\definecolor{jgray}{RGB}{10, 10, 10}
\definecolor{silver}{RGB}{192, 192, 192}

\definecolor{wxjiao}{RGB}{18, 141, 21}

\title{
Is ChatGPT A Good Translator? Yes With GPT-4 As The Engine
}

\author{
Wenxiang Jiao\thanks{~~Corresponding author.} \enspace Wenxuan Wang \enspace Jen-tse Huang \enspace Xing Wang \\[0.5ex]
\bf Shuming Shi \enspace Zhaopeng Tu \\[1ex]
Tencent AI Lab \\
{\asciifamily \normalsize \tt \{joelwxjiao\}@tencent.com} \\
}

\begin{document}
\maketitle
\begin{abstract}
  This report provides a preliminary evaluation of ChatGPT for machine translation, including translation prompt, multilingual translation, and translation robustness. 
  We adopt the prompts advised by ChatGPT to trigger its translation ability and find that the candidate prompts generally work well with minor performance differences.
  By evaluating on a number of benchmark test sets, we find that ChatGPT performs competitively with commercial translation products (e.g., Google Translate) on high-resource European languages but lags behind significantly on low-resource or distant languages.
  As for the translation robustness, ChatGPT does not perform as well as the commercial systems on biomedical abstracts or Reddit comments but exhibits good results on spoken language. 
  Further, we explore an interesting strategy named \textbf{pivot prompting} for distant languages, which asks ChatGPT to translate the source sentence into a high-resource pivot language before into the target language, improving the translation performance noticeably.
  With the launch of the GPT-4 engine, the translation performance of ChatGPT is significantly boosted, becoming comparable to commercial translation products, even for distant languages.
  Human analysis on Google Translate and ChatGPT suggests that ChatGPT with GPT-3.5 tends to generate more hallucinations and mis-translation errors while that with GPT-4 makes the least errors.
  In other words, ChatGPT has already become a good translator.
  Please refer to our Github project for more details: \url{https://github.com/wxjiao/Is-ChatGPT-A-Good-Translator}.
\end{abstract}

\section{Introduction}
\label{sec:introduction}

ChatGPT\footnote{\url{https://chat.openai.com}} is an intelligent chatting machine developed by OpenAI upon the InstructGPT~\cite{ouyang2022InstructGPT}, which is trained to follow an instruction in a prompt and provide a detailed response. According to the official statement, ChatGPT is able to answer followup questions, admit its mistakes, challenge incorrect premises, and reject inappropriate requests due to the dialogue format. It integrates various abilities of natural language processing, including question answering, storytelling, logic reasoning, code debugging, machine translation, and so on.
We are particularly interested in how ChatGPT performs for machine translation tasks, especially the gap between ChatGPT and commercial translation products (e.g., Google Translate, DeepL Translate).

\begin{figure}[t]
    \centering
    \includegraphics[width=\linewidth]{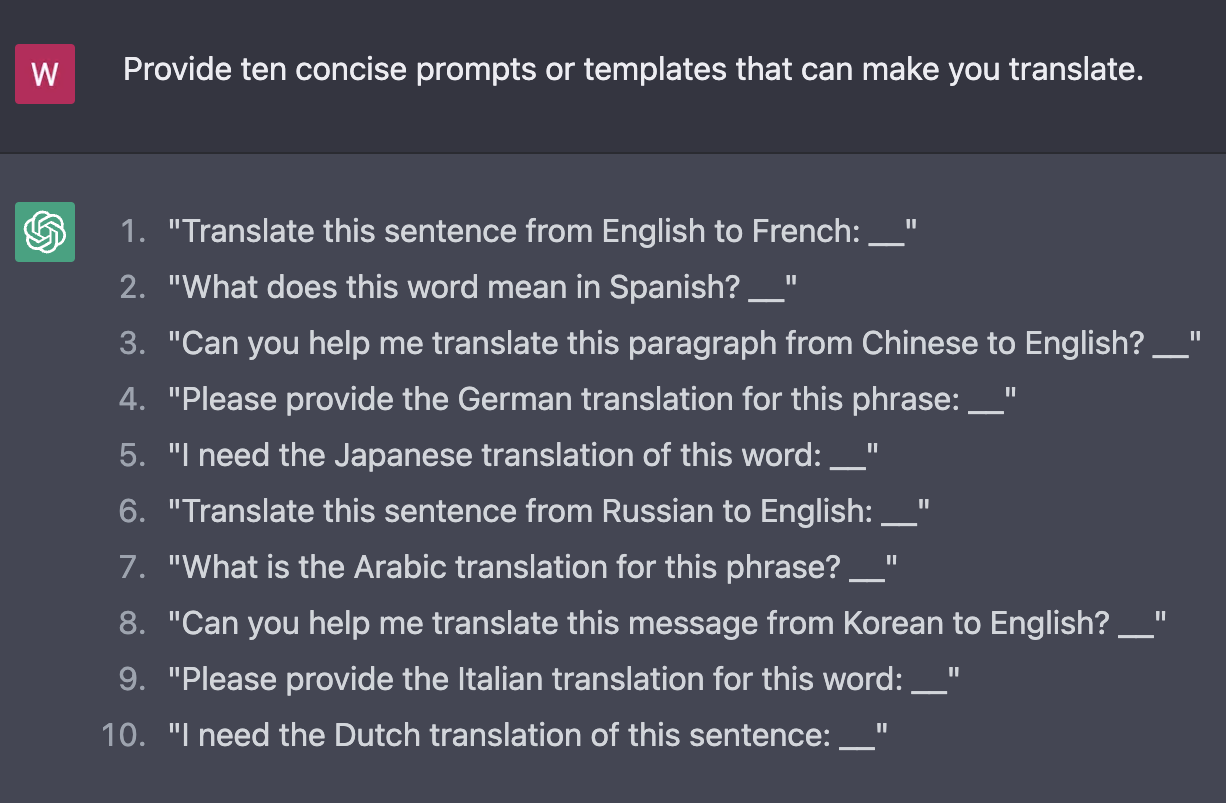}
    \caption{Prompts advised by ChatGPT for machine translation (Date: 2022.12.16). }
    \label{fig:prompts-chatgpt}
\end{figure}

In this report, we provide a preliminary study of ChatGPT on machine translation, which to our best knowledge is also the first one since the release of ChatGPT. Specifically, we focus on three aspects:
\begin{itemize}[leftmargin=10pt]
    \item \textbf{Translation Prompt}: ChatGPT is essentially a large language model, which needs prompts as guidance to trigger its translation ability. The style of prompts may affect the quality of translation outputs. For example, how to mention the source or target language information matters in multilingual machine translation models, which is usually solved by attaching language tokens~\cite{Johnson:2017:TACL,fan2021beyond}.
    \item \textbf{Multilingual Translation}: ChatGPT is a single model handling various NLP tasks and covering different languages, which can be considered a unified multilingual machine translation model. Thus, we are curious about how ChatGPT performs on different language pairs considering both the resource difference (e.g., high vs. low) and language family (e.g., European vs. Asian).
    \item \textbf{Translation Robustness}: ChatGPT is developed upon GPT3, which was trained on large-scale datasets that cover various domains. Therefore, we wonder if it can perform robustly well on domain-specific or even noisy sentences.
\end{itemize}

To trigger the translation ability of ChatGPT, we ask ChatGPT itself for advice and obtain three candidate translation prompts. By evaluating on the Chinese$\Rightarrow$English translation task, we find that the candidate prompts generally work well and show minor performance differences. Nevertheless, we adopt the best-performing prompt for the rest parts of the study. 
By evaluating the translation among four selected languages on the Flores-101 test sets, we find that ChatGPT performs competitively with commercial translation products (e.g., Google Translate) on high-resource European languages but lags behind significantly on low-resource or distant languages.
As for the translation robustness, results on three robustness sets suggest that ChatGPT does not perform as well as the commercial systems on biomedical abstracts or Reddit comments but exhibits good results on spoken language.

Further, we have a discussion on how to improve ChatGPT for machine translation.
On one hand, we explore an interesting strategy named \textbf{pivot prompting} for distant languages, which asks ChatGPT to translate the source sentence into a high-resource pivot language before into the target language, improving the translation performance noticeably. 
On the other hand, with an improved engine GPT-4~\cite{openai2023gpt4}\footnote{\url{https://openai.com/research/gpt-4}} launched on March 15, 2023, we re-evaluate the translation ability of ChatGPT and observe a significant boost of performance. The translation performance of ChatGPT becomes comparable to commercial translation products, even for distant languages.
Extensive analysis on Google Translate and ChatGPT suggests that ChatGPT with GPT-3.5 tend to generate more hallucinations and more mis-translation errors while that with GPT-4 makes the least errors.
In other words, \textbf{ChatGPT has already become a good translator with GPT-4 as the engine!}

\begin{table}[t!]
\setlength{\tabcolsep}{2pt}
    \centering
    \caption{Information of adopted test sets.}
    \resizebox{1.0\columnwidth}{!}{
    \begin{tabular}{l ccr}
    \toprule
     \bf  Test Set & \bf Direction & \bf Domain & \bf Size \\
     \midrule
     Flores-101 & Any & General & 1012 \\
     \hline
     WMT19 Bio & De$\Rightarrow$En & Biomedical & 373 \\
     \hdashline
     \multirow{2}{*}{WMT20 Rob2} & En$\Rightarrow$Ja & \multirow{2}{*}{Reddit} & 1376 \\
      & Ja$\Rightarrow$En &  & 997 \\
     \hdashline
     WMT20 Rob3 & De$\Rightarrow$En & Common Voice & 5609 \\
    \bottomrule
    \end{tabular}
    }
    \label{tab:info-test-sets}
\end{table}

\section{ChatGPT for Machine Translation}

\subsection{Evaluation Setting}
We provide a brief introduction of the evaluation setting, which mainly includes the compared baselines and test data.
\paragraph{Baselines.}
We compare ChatGPT with three commercial translation products, namely, Google Translate\footnote{\url{https://translate.google.com}}, DeepL Translate\footnote{\url{https://www.deepl.com/translator}}, and Tencent TranSmart\footnote{\url{https://transmart.qq.com/zh-CN/index}}. So far, the three commercial systems support translation in 133, 29, and 16 languages, respectively.
By default, the results in this report come from the ChatGPT version on \colorbox{red!20}{2022.12.16}. For new results, we will mark the updated version information correspondingly.

\paragraph{Data.}
For multilingual translation, we evaluate the above translation systems on the Flores-101~\cite{goyal2021flores}\footnote{\url{https://github.com/facebookresearch/flores}} test sets, which consists of 1012 sentences translated into 101 languages. To test the translation robustness, we adopt the test set of WMT19 Biomedical Translation Task~\cite[i.e., Bio]{bawden2019findings} and the set2 and set3 of WMT20 Robustness Task~\cite[i.e., Rob2 and Rob3]{specia2020findings}. 
We obtain the first two test sets through SacreBLEU and the third pre-processed by \newcite{wang2021tencent}\footnote{\url{https://github.com/hsing-wang/WMT2020_BioMedical/tree/master/Bio-18-19-testset}}.
Table~\ref{tab:info-test-sets} lists the information of these test sets.
Since this empirical study was conducted upon the very early release of ChatGPT, we were only able to access it through webpage, which cannot respond to large batches. As a result, obtaining the translation results from ChatGPT is time-consuming.
Therefore, we randomly sample 50 sentences from each set for evaluation.

\paragraph{Metric.}
We adopt the mostly used BLEU score~\cite{papineni2002bleu} as our primary metric and also report ChrF++~\cite{popovic2017chrf++} and TER~\cite{snover2006ter} in some cases. These three metrics are all supported by SacreBLEU~\cite{post2018sacrebleu}\footnote{\url{https://github.com/mjpost/sacrebleu}}.

\begin{table}[t!]
\centering
\caption{Candidate translation prompts.}
\begin{tabular}{c p{6cm}}
\toprule
& \multicolumn{1}{c}{\bf Translation Prompt} \\
\midrule
\textsc{Tp1} & \texttt{Translate these sentences from [SRC] to [TGT]: } \\
\textsc{Tp2} & \texttt{Answer with no quotes. What do these sentences mean in [TGT]?} \\
\textsc{Tp3} & \texttt{Please provide the [TGT] translation for these sentences:} \\
\bottomrule
\end{tabular}
\label{tab:prompts-candidate}
\end{table}

\subsection{Translation Prompts}
\label{sec:translation-prompts}

To design the prompts for triggering the machine translation ability of ChatGPT, we seek inspiration from ChatGPT by asking it for advice. Specifically, we ask ChatGPT with the following prompt:
\begin{quote}
\texttt{Provide ten concise prompts or templates that can make you translate.}
\end{quote}
and obtain the results as shown in Figure~\ref{fig:prompts-chatgpt}. The generated prompts look reasonable but share similar formats. Thus, we summarize them into three candidate prompts as shown in Table~\ref{tab:prompts-candidate}, where \texttt{[SRC]} and \texttt{[TGT]} represent the source and target languages of translation. Note that we add an extra command into \textsc{Tp2} to ask ChatGPT not to generate double quotes around the translation, which often occurs with the original format. Nevertheless, it is still unstable such that sentences in a batch (in multiple lines) are translated into a single line occasionally. 

We compare the three different candidate prompts on the Chinese-to-English~(Zh$\Rightarrow$En) translation task with the test set from Flores-101. 
Table~\ref{tab:bleu-prompts-chatgpt} shows the results of ChatGPT and three commercial systems. 
While ChatGPT provides reasonably good translations, it still lags behind the baselines by at least 5.0 BLEU points. Concerning the three candidate prompts, \textsc{Tp3} performs the best in terms of all the three metrics. Thus, we use \textsc{Tp3} throughout this report by default.

\subsection{Multilingual Translation}
\label{sec:multilingual-translation}

\begin{table}[t]
\setlength{\tabcolsep}{5pt}
\centering
\caption{
Comparison of different prompts for ChatGPT to perform Chinese-to-English~(Zh$\Rightarrow$En) translation.}
\begin{threeparttable}
\begin{tabular}{l cccccc}
\toprule
\bf System & \bf BLEU$^\uparrow$ & \bf ChrF++$^\uparrow$ & \bf TER$^\downarrow$ \\
\midrule
Google  & 31.66 & 57.09 & 56.21  \\
DeepL & 31.22 & 56.74 & 57.84  \\
Tencent & 29.69 & 56.24 & 57.16 \\
\hline
ChatGPT w/ \textsc{Tp1} & 23.25 & 53.07 & 66.03 \\
ChatGPT w/ \textsc{Tp2} & 24.54 & 53.05 & 63.79 \\
ChatGPT w/ \textsc{Tp3} & \bf 24.73 & \bf 53.71 & \bf 62.84 \\
\bottomrule
\end{tabular}
\end{threeparttable}
\label{tab:bleu-prompts-chatgpt}
\end{table}

\begin{table*}[t!]
\centering
\caption{
Performance of ChatGPT for multilingual translation. }
\begin{tabular}{l llllll}
\toprule
\multirow{2}{*}{\bf System}
& \multicolumn{2}{c}{\bf De-En}
& \multicolumn{2}{c}{\bf Ro-En}
& \multicolumn{2}{c}{\bf Zh-En}\\
\cmidrule(lr){2-3}\cmidrule(lr){4-5}\cmidrule(lr){6-7}
& \multicolumn{1}{c}{$\Rightarrow$} & \multicolumn{1}{c}{$\Leftarrow$} & \multicolumn{1}{c}{$\Rightarrow$} & \multicolumn{1}{c}{$\Leftarrow$} & \multicolumn{1}{c}{$\Rightarrow$} & \multicolumn{1}{c}{$\Leftarrow$} \\
\midrule
Google  & 45.04 & 41.16 & 50.12 & 46.03 & 31.66 & 43.58 \\
DeepL & 49.23\tiny{(+9.3\%)} & 41.46\tiny{(+0.7\%)} & 50.61\tiny{(+0.9\%)} & 48.39\tiny{(+5.1\%)} & 31.22\tiny{(-1.3\%)} & 44.31\tiny{(+1.6\%)} \\
Tencent & n/a & n/a & n/a & n/a & 29.69\tiny{(-6.2\%)} & 46.06\tiny{(+5.6\%)} \\
\hline
ChatGPT & 43.71\tiny{(-2.9\%)} & 38.87\tiny{(-5.5\%)} & 44.95\tiny{(-10.3\%)} & 24.85\tiny{(-46.0\%)} & 24.73\tiny{(-21.8\%)} & 38.27\tiny{(-12.1\%)} \\
\midrule
\toprule
\multirow{2}{*}{\bf System}
& \multicolumn{2}{c}{\bf De-Zh}
& \multicolumn{2}{c}{\bf Ro-Zh}
& \multicolumn{2}{c}{\bf De-Ro}\\
\cmidrule(lr){2-3}\cmidrule(lr){4-5}\cmidrule(lr){6-7}
& \multicolumn{1}{c}{$\Rightarrow$} & \multicolumn{1}{c}{$\Leftarrow$} & \multicolumn{1}{c}{$\Rightarrow$} & \multicolumn{1}{c}{$\Leftarrow$} & \multicolumn{1}{c}{$\Rightarrow$} & \multicolumn{1}{c}{$\Leftarrow$} \\
\midrule
Google  & 38.71 & 21.68 & 39.05 & 25.59 & 33.31 & 32.27 \\
DeepL & 40.46\tiny{(+4.5\%)} & 22.82\tiny{(+5.2\%)} & 38.95\tiny{(-0.2\%)} & 25.39\tiny{(-0.7\%)} & 35.19\tiny{(+5.6\%)} & 34.27\tiny{(+6.1\%)} \\
Tencent & 40.66\tiny{(+5.0\%)} & 19.44\tiny{(-10.3\%)} & n/a & n/a & n/a & n/a \\
\hline
ChatGPT & 34.46\tiny{(-10.9\%)} & 19.80\tiny{(-8.6\%)} & 30.84\tiny{(-21.0\%)} & 19.17\tiny{(-25.0\%)} & 33.38\tiny{(+0.2\%)} & 29.89\tiny{(-7.3\%)} \\
\bottomrule
\end{tabular}
\label{tab:bleu-multilingual-chatgpt}
\end{table*}

We select four languages to evaluate the capability of ChatGPT in multilingual translation, including German (De), English (En), Romanian (Ro), and Chinese (Zh), which are commonly adopted in both research~\citep{wang2022understanding,jiao2021self,jiao2022exploiting} and competitions~\citep{bojar2016wmt,farhad2021wmt}. 
The first three languages come from the same family with Latin scripts while the last is from another family with Chinese scripts~\cite{fan2021beyond}. 
We test the translation performance between any two languages, which involves 12 directions in total. 
For clarity and comparison, we report the BLEU scores and the improvement or drop of performance (i.e., +/-) relative to Google Translate.
Table~\ref{tab:bleu-multilingual-chatgpt} presents the results.

\paragraph{Resource Difference.} 
We consider the resource difference of languages in the same family.
In machine translation, German$\Leftrightarrow$English translation is usually regarded as a high-resource task supported by over ten million sentence pairs~\cite{farhad2021wmt} while Romanian$\Leftrightarrow$English translation is supported by much less data~\cite{bojar2016wmt}. 
This resource difference can also be indicated by the data statistics\footnote{\url{https://github.com/openai/gpt-3/tree/master/dataset_statistics}} of GPT-3~\cite{brown2020gpt3}, although we do not know the data information of ChatGPT.
As shown in Table~\ref{tab:bleu-multilingual-chatgpt}, ChatGPT performs competitively with Google Translate and DeepL Translate for both German$\Rightarrow$English and English$\Rightarrow$German translations. However, it lags behind them significantly on Romanian$\Rightarrow$English and English$\Rightarrow$Romanian. Specifically, ChatGPT obtains a BLEU score on English$\Rightarrow$Romanian that is 46.4\% lower than Google Translate and the value is 10.3\% on Romanian$\Rightarrow$English.
We speculate that the huge resource difference of monolingual data between English and Romanian limits the language modeling capability of Romanian, which partially explains the poor performance on English$\Rightarrow$Romanian. On the contrary, Romanian$\Rightarrow$English can benefit from the strong language modeling capability of English such that the resource gap of parallel data can be somewhat compensated.

\paragraph{Language Family.} 
We also take the impact of language families into account. In machine translation, translating between different language families is often considered harder than that within the same language family, due to the different cultures and writing scripts.
By comparing German$\Leftrightarrow$English with Chinese$\Leftrightarrow$English or German$\Leftrightarrow$Chinese translation, we find that the gap between ChatGPT and the commercial systems becomes larger. We attribute to the better knowledge transfer within the same family (i.e., from English to German) than between different families (e.g., from English to Chinese). For language pairs that are both low-resource and from different families (e.g., Romanian$\Leftrightarrow$Chinese), the performance gap can be further enlarged~\cite{wang2022uncertainty}.
Since ChatGPT handles different tasks in one model, low-resource translation tasks not only compete with high-resource translation tasks~\cite{jiao2022tencent}, but also with other NLP tasks for the model capacity, which explains their poor performance.

\subsection{Translation Robustness}
\label{sec:translation-robustness}

We further evaluate the translation robustness of ChatGPT on the WMT19 Bio and WMT20 Rob2 and Rob3 test sets, which introduce the impact of domain bias and potentially noisy data.
For example, WMT19 Bio test set is composed of Medline abstracts, which require domain-specific knowledge to handle the terminologies.
WMT20 Rob2 are comments from the social media website \url{reddit.com} that could contain various errors, including spelling/typographical errors, word omission/insertion/repetition, grammatical errors, spoken languages, Internet slang, and so on~\cite{michel2018mtnt}.

\begin{table}[t]
\setlength{\tabcolsep}{2pt}
\centering
\caption{
Performance of ChatGPT for translation robustness on domain-specific or noisy test data. }
\resizebox{1.0\columnwidth}{!}{
\begin{tabular}{l cccccccc}
\toprule
\multirow{2}{*}{\bf System}
& \multicolumn{1}{c}{\bf W19 Bio}
& \multicolumn{2}{c}{\bf W20 Rob2}
& \multicolumn{1}{c}{\bf W20 Rob3}\\
\cmidrule(lr){2-2}\cmidrule(lr){3-4}\cmidrule(lr){5-5}
& De$\Rightarrow$En & En$\Rightarrow$Ja & Ja$\Rightarrow$En & De$\Rightarrow$En \\
\midrule
Google  & 37.83 & 29.72 & 19.21 & 42.91 \\
DeepL & 37.13 & 26.25 & 19.83 & 41.29 \\
\hline
ChatGPT & 33.22 & 22.36 & 18.34 & 44.59 \\
\bottomrule
\end{tabular}
}
\label{tab:bleu-robust-chatgpt}
\end{table}

Table~\ref{tab:bleu-robust-chatgpt} lists the BLEU scores. 
Obviously, ChatGPT does not perform as well as Google Translate or DeepL Translate on the WMT19 Bio and WMT2 Rob2 test sets.
The reason may be that commercial translation systems like Google Translate often need to continuously improve their ability to translate domain-specific (e.g. biomedical) or noisy sentences, since they are real-world applications that require better generalization performance over out-of-distribution data.
However, these may not be done in ChatGPT.

An interesting finding is that ChatGPT outperforms Google Translate and DeepL Translate significantly on WMT20 Rob3 test set that contains a crowdsourced speech recognition corpus.
It suggests that ChatGPT, which is essentially an artificial intelligent chatting machine, is capable of generating more natural spoken languages than these commercial translation systems. We provide some examples in Table~\ref{tab:cases-robust-chatgpt}.

\begin{table}[t]
\centering
\caption{Examples from WMT20 Robust Set3.}
\resizebox{1.0\columnwidth}{!}{
\begin{tabular}{l p{5cm}}
\toprule
& \multicolumn{1}{c}{\bf Example} \\
\midrule
\textsc{Src} & Haben wir noch Nudeln? \\
\textsc{Ref} & Do we still have noodles? \\
\hdashline
Google & Do we still have pasta? \\
DeepL & Do we have any noodles left? \\
ChatGPT & Do we still have noodles? \\
\hline
\textsc{Src} & Tatsächlich ist der zu häufige Gebrauch von Seife schlecht für die Haut. \\
\textsc{Ref} & Actually, very frequent usage of soap is bad for the skin. \\
\hdashline
Google & In fact, using soap too often is bad for your skin. \\
DeepL & In fact, using soap too often is bad for the skin. \\
ChatGPT & In fact, the frequent use of soap is bad for the skin. \\
\bottomrule
\end{tabular}
}
\label{tab:cases-robust-chatgpt}
\end{table}

\section{Improving ChatGPT for MT}

As presented above, ChatGPT can match the performance of commercial translation systems on high-resource language pairs, but still struggles on low-resource ones, especially those distant languages. Then, one question arises: 
\begin{quote}
    \it How can we improve ChatGPT for MT?
\end{quote}

\subsection{Pivot Prompting}
\label{sec:pivot-prompting}

\begin{figure}[t!]
    \centering
    \includegraphics[width=\linewidth]{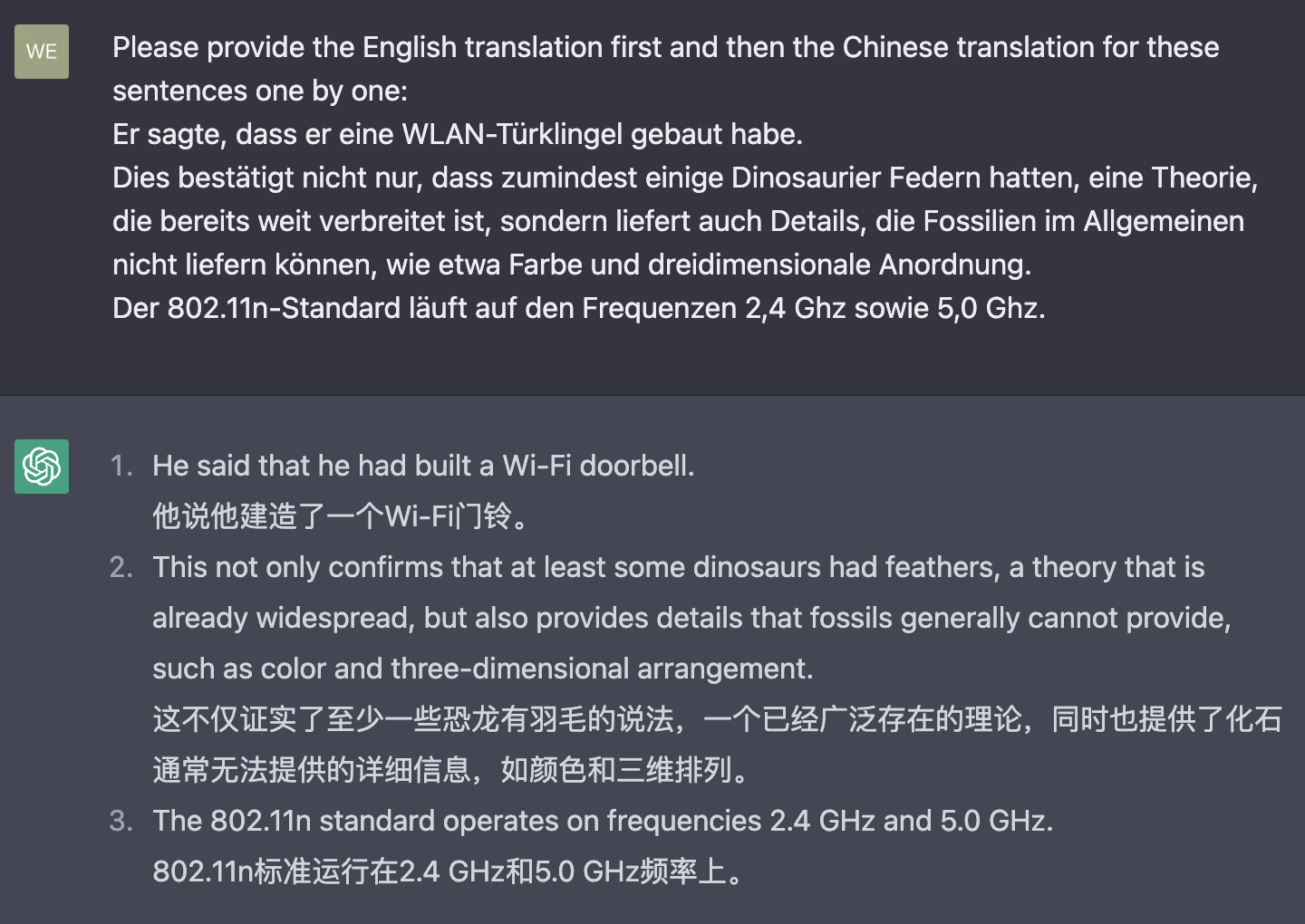}
    \caption{Translation results by ChatGPT with pivot prompting (Date: \colorbox{red!20}{2023.01.31}). }
    \label{fig:prompts-pivot-chatgpt}
\end{figure}

The first way to improve ChatGPT for MT is to exploit the potential of ChatGPT in other tasks to assist the target task.
Here, we explore an interesting strategy named \textit{Pivot Prompting} to improve the translation quality between distant languages. 
Rather than the direct translation between source and target languages, we ask ChatGPT to translate the source sentence into a high-resource pivot language (i.e., English by default) first and then into the target language.
Accordingly, we adjust the \textsc{Tp3} prompt as below:
\begin{quote}
\texttt{Please provide the [PIV] translation first and then the [TGT]
translation for these sentences one by one:}
\end{quote}
where \texttt{[PIV]} denotes the pivot language. As a large language model, ChatGPT will naturally condition on both the prompt and the translation result in the pivot language to generate the translation into the target language. Figure~\ref{fig:prompts-pivot-chatgpt} shows an example when using pivot prompting.

\begin{table}[t]
\setlength{\tabcolsep}{3pt}
\centering
\caption{Performance of ChatGPT with pivot prompting. New results are obtained from the updated ChatGPT version on 2023.01.31. LR: length ratio.}
\begin{tabular}{l cccccc}
\toprule
\multirow{2}{*}{\bf System} & \multicolumn{2}{c}{\bf De$\Rightarrow$Zh} & \multicolumn{2}{c}{\bf Ro$\Rightarrow$Zh} \\
\cmidrule(lr){2-3}\cmidrule(lr){4-5}
& BLEU & LR & BLEU & LR \\
\midrule
Google  & 38.71 & 0.94 & 39.05 & 0.95 \\
DeepL & 40.46 & 0.98 & 38.95 & 0.99 \\
\hline
ChatGPT (Direct) & 34.46 & 0.97 & 30.84 & 0.91 \\
ChatGPT (Direct$_{\rm new}$) & 30.76 & 0.92 & 27.51 & 0.93 \\
ChatGPT (Pivot$_{\rm new}$) & 34.68 & 0.95 & 34.19 & 0.98 \\
\bottomrule
\end{tabular}
\label{tab:bleu-pivot-prompting}
\end{table}

\begin{table*}[t!]
\centering
\caption{Performance of GPT-4 (Date: \colorbox{red!20}{2023.03.15}) for multilingual translation.}
\begin{tabular}{l cccccc}
\toprule
\bf System & \bf De$\Rightarrow$En & \bf En$\Rightarrow$De & \bf Zh$\Rightarrow$En & \bf En$\Rightarrow$Zh & \bf De$\Rightarrow$Zh & \bf Ro$\Rightarrow$Zh \\
\midrule
Google  & 45.04 & 41.16 & 31.66 & 43.58 & 38.71 & 39.05 \\
DeepL & 49.23 & 41.46 & 31.22 & 44.31 & 40.46 & 38.95 \\
Tencent & n/a & n/a & 29.69 & 46.06 & 40.66 & n/a \\
\hline
ChatGPT (Direct) & 43.71 & 38.87 & 24.73 & 38.27 & 34.46 & 30.84 \\
ChatGPT (Direct$_{\rm new}$) & n/a & n/a & n/a & n/a & 30.76 & 27.51 \\
ChatGPT (Pivot$_{\rm new}$) & n/a & n/a & n/a & n/a & 34.68 & 34.19 \\
\hline
\bf GPT-4 & 46.00 & 45.73 & 28.50 & 42.50 & 38.16 & 37.84 \\
\bottomrule
\end{tabular}
\label{tab:bleu-GPT4}
\end{table*}

There are several advantages of pivot prompting:
\begin{itemize}[leftmargin=10pt]
    \item \textbf{Knowledge Transfer}: While parallel data between two distant languages is often scarce~\cite{fan2021beyond,wang2022uncertainty}, the parallel data between them and the pivot language can be relatively considerable, which is expected to learn better translation ability for source-pivot and pivot-target directions than that for the source-target direction. Thus, pivot prompting will potentially transfer the knowledge of the high-resource pivot language to the low-resource target languages~\cite{zoph2016transfer,aji2020neural,li2022consisttl,he2022tencent}.
    \item \textbf{Convenience}: Essentially, pivot prompting is similar to the pivot translation technique in previous studies~\cite{cheng2016pivot} but is more convenient for ChatGPT. For the commonly adopted multilingual sequence-to-sequence translation models~\cite{fan2021beyond}, pivot translation requires two steps: (1) Input the source sentence and translate it into the pivot language; (2) Input the translation results in pivot language and translate it into the target language. 
    In contrast, ChatGPT can identify both the \texttt{[PIV]} and \texttt{[TGT]} languages and translate the source sentence into the two languages sequentially~(see Figure~\ref{fig:prompts-pivot-chatgpt}), which requires only one step operation.
\end{itemize}

Table~\ref{tab:bleu-pivot-prompting} presents our results in BLUE score and length ratio of translation results over references. We obtain the translation results by using \textsc{Tp3}~(i.e., Direct) and pivot prompting~(i.e., Pivot) through English~(i.e., source-to-English-to-target), respectively. As seen, the latest update for ChatGPT seems to harm the translation quality for German$\Rightarrow$Chinese and Romanian$\Rightarrow$Chinese translations, compared with the previous version we used (i.e., Direct$_{\rm new}$ vs. Direct).
Nevertheless, pivot prompting can significantly improve the translation performance by nearly 3.9 and 6.6 BLEU points for German$\Rightarrow$Chinese and Romanian$\Rightarrow$Chinese translations, respectively, which demonstrates its effectiveness.
By inspecting the translation results, we find that direct translation with \textsc{Tp3} will under-translate some tokens in source sentences, which can be noticeably fixed by pivot prompting. This can be reflected by the length ratio results.
Note that, while pivot prompting is convenient for ChatGPT, how to further accelerate the inference process is still an important research question as we need to generate longer sentences.

\subsection{GPT-4 as the Engine}

Another way to improve ChatGPT for MT is to improve its engine. Unsurprisingly, OpenAI released GPT-4~\cite{openai2023gpt4} on \colorbox{red!20}{March 15, 2023}, which exhibits all-around stronger capabilities than the GPT-3.5 model behind ChatGPT.
Therefore, we re-evaluate the performance for four translation directions. As shown in Table~\ref{tab:bleu-GPT4}, GPT-4 boosts the performance over ChatGPT significantly on all the four directions, bringing the BLEU scores to the level of top commercial translation systems. Note that these results only come from zero-shot settings. With modern techniques like in-context learning with demonstrations~\cite{brown2020gpt3,agrawal2022context}, the translation performance could be further improved. In other words, \textbf{GPT-4 has already become a good translator!}

\section{Analysis}
\label{sec:analysis}

Here we conduct some analyses on the translation outputs for a deeper understanding in ChatGPT.
By default, we analyze the outputs of Google, ChatGPT, and GPT-4 on Zh$\Rightarrow$En translation for all the 50 test examples.

\subsection{Automatic Analysis}

We follow previous studies~\cite{jiao2021self,wang2022understanding} to analyze the translation outputs using automatic tools, i.e., \texttt{compare-mt}\footnote{\url{https://github.com/neulab/compare-mt}}, at both word level and sentence level.

\paragraph{Word Frequency.}
Essentially, ChatGPT is a large language model that has been trained on a variety of corpora, covering different domains. It could be beneficial to the translation of low-frequency words in the test sets.
Specifically, we divide the target words into three categories based on their frequency and calculate the accuracy of word prediction. Table~\ref{tab:analysis-freq} shows the F-measure results.
Unexpectedly, ChatGPT turns out to perform the worst on low-frequency words (i.e., $< 2$), which we attribute to the immature translation ability of ChatGPT.
What's interesting is that GPT-4 mainly addresses this shortcoming for ChatGPT with little improvement to higher-frequency words.

\begin{table}[t]
\centering
\caption{F-measure of target word prediction with respect to frequency bucket.}
\begin{tabular}{c cccccc}
\toprule
\bf Freq & \bf Google & \bf ChatGPT & \bf GPT-4 \\
\midrule
$< 2$ & 48.0 & 43.7 & 47.1 \\
$[2, 10)$ & 59.0 & 57.6 & 56.7 \\
$\geq 10$ & 71.6 & 70.5 & 70.1 \\
\bottomrule
\end{tabular}
\label{tab:analysis-freq}
\end{table}

\begin{table}[t]
\centering
\caption{Translation performance (i.e., BLEU) with respect to length bucket of target sentences.}
\begin{tabular}{c cccccc}
\toprule
\bf Length & \bf Google & \bf ChatGPT & \bf GPT-4 \\
\midrule
$< 15$ & 34.2 & 15.4 & 26.1 \\
$[15, 30)$ & 26.6 & 21.4 & 24.3 \\
$\geq 30$ & 23.2 & 16.0 & 19.4 \\
\bottomrule
\end{tabular}
\label{tab:analysis-length}
\end{table}

\paragraph{Sentence Length.}
ChatGPT is also trained for various text generation tasks, which usually do not require strict length constraints of generated sentences as machine translation.
Therefore, we are curious about how sensitive the translation performance is to the sentence length. 
We divide the target sentences into three categories based on the sentence length, of which the average value is 23.2 tokens.
Table~\ref{tab:analysis-length} shows the results. As seen, ChatGPT performs the worst on short sentences (i.e., $< 15$), with 18.8 BLEU points lower than Google Translate. One observation is that when translating terminologies, e.g., \begin{CJK}{UTF8}{gkai}美国公共广播公司\end{CJK}, ChatGPT tends to output the full names~(i.e., American Public Broadcasting System) while Google Translate and the reference use the abbreviations~(i.e., PBS).
As a result, the precision of word prediction will be reduced noticeably, so will BLEU score~\cite{papineni2002bleu}, especially for short sentences.
GPT-4 can predict the abbreviations properly sometimes, which gives a better translation performance.

\begin{table}[t]
\centering
\caption{Number of translation errors annotated by human.}
\begin{tabular}{c cccccc}
\toprule
\bf Error & \bf Google & \bf ChatGPT & \bf GPT-4 \\
\midrule
Und-Trans & 9 & 5 & 5 \\
Ove-Trans & 6 & 8 & 1  \\
Mis-Trans & 16 & 23	& 7 \\
\bottomrule
\end{tabular}
\label{tab:analysis-errors}
\end{table}

\begin{table}[t]
\centering
\caption{Human rankings of the translation outputs.}
\begin{tabular}{c cccccc}
\toprule
\bf Rank & \bf Google & \bf ChatGPT & \bf GPT-4 \\
\midrule
1 & 20 & 11 & 32  \\
2 & 14 & 19 & 13 \\
3 & 16 & 20 & 5  \\
\bottomrule
\end{tabular}
\label{tab:analysis-ranking}
\end{table}

\subsection{Human Analysis}

In addition to the automatic analysis, we also inspect the translation outputs manually. We ask three annotators to identify the errors in the translation outputs~\cite{wang2022understanding}, including under-translation~(i.e., Und-Trans), over-translation~(i.e., Ove-Trans), and mis-translation~(i.e., Mis-Trans). Based on the translation errors, the annotators rank the translation outputs of Google, ChatGPT and GPT-4 accordingly, with 1 as the best system and 3 as the worst.
For translation outputs that are really hard to distinguish, we allow the same ranking (e.g., 1-1-1, 1-1-2 or 1-2-2).
To eliminate subjective bias, we do not present the system information of each translation output to the annotators, and the three translation outputs for each test example are also shuffled randomly.  

Table~\ref{tab:analysis-errors} shows the results of translation errors.
Generally, ChatGPT makes more over-translation errors and mis-translation errors than Google Translate, but slightly less under-translation errors. It suggests that ChatGPT is more likely to generate hallucinations.
In contrast, GPT-4 makes the least errors across the three error classes, which demonstrates the best translation performance.
This is also confirmed by the ranking results in Table~\ref{tab:analysis-ranking}, such that GPT-4 is ranked the best (i.e., 1) for 32 times out of 50 test examples, followed by Google Translate and ChatGPT.
However, the BLEU score of GPT-4 is still lower than that of Google Translate (i.e., 28.50 vs. 31.66 in Table~\ref{tab:bleu-GPT4}), which indicates that GPT-4 may generate more diverse translations with different lexical choices from the references.

\begin{table*}[t!]
\fontsize{9}{11}\selectfont
\setlength{\tabcolsep}{3pt}
\centering
\caption{Examples from Flores Zh$\Rightarrow$En test set.}
\begin{tabular}{l p{14.2cm}}
\toprule
& \multicolumn{1}{c}{\bf Example} \\
\midrule
\textsc{Src} & \begin{CJK}{UTF8}{gkai} \colorbox{blue!20}{狂风}、冰雹、\colorbox{blue!20}{过量降水}、山火，还有雷雨、龙卷风、水龙卷、气旋，都是极端天气的表现和影响。 \end{CJK} \\
\textsc{Ref} & High winds, hail, excessive precipitation, and wildfires are forms and effects of severe weather, as are thunderstorms, tornadoes, waterspouts, and cyclones. \\[0.5ex]
\hdashline
Google & \colorbox{red!20}{Strong winds}, hail, \colorbox{red!20}{excessive precipitation}, wildfires, as well as thunderstorms, tornadoes, waterspouts, and cyclones are all manifestations and effects of extreme weather. \\
ChatGPT & \colorbox{red!20}{Extreme weather conditions such as strong winds}, hail, \colorbox{red!20}{excessive rainfall}, wildfires, thunderstorms, tornadoes, waterspouts, and cyclones are all expressions and impacts of extreme weather. \\
GPT-4 & \colorbox{red!20}{High winds}, hail, \colorbox{red!20}{excessive precipitation}, wildfires, as well as thunderstorms, tornadoes, waterspouts, and cyclones are all manifestations and impacts of extreme weather. \\
\midrule
\textsc{Src} & \begin{CJK}{UTF8}{gkai} 然而，\colorbox{blue!20}{广泛耐药结核病}在整个结核病患者群体中所占的比例似乎仍然很低；在南非，在任意特定时刻，总计 33 万人中只有 6000 人感染。 \end{CJK} \\
\textsc{Ref} & However, the percentage of \colorbox{red!20}{XDR-TB} in the entire group of people with tuberculosis still seems to be low; 6,000 of the total 330,000 people infected at any particular moment in South Africa. \\[0.5ex]
\hdashline
Google & However, the proportion of \colorbox{red!20}{XDR-TB} in the overall TB patient population still appears to be low; in South Africa, only 6000 of a total of 330,000 people are infected at any given moment. \\
ChatGPT & However, the proportion of \colorbox{red!20}{extensively drug-resistant tuberculosis} in the overall tuberculosis patient population seems to remain low; in South Africa, at any given time, only 6,000 out of a total of 330,000 people are infected. \\
GPT-4 & However, the proportion of \colorbox{red!20}{extensively drug-resistant tuberculosis} among the entire tuberculosis patient population still seems to be quite low; in South Africa, at any given time, only 6,000 out of a total of 330,000 people are infected. \\
\midrule
\textsc{Src} & \begin{CJK}{UTF8}{gkai}  这是亨利·路易斯·盖茨 (Henry Louis Gates) 在\colorbox{blue!20}{美国公共广播公司}的《非洲世界的奇迹》特别节目中的重要一站。 \end{CJK} \\
\textsc{Ref} & It was one of the major stops during Henry Louis Gates' \colorbox{red!20}{PBS} special Wonders of the African World. \\[0.5ex]
\hdashline
Google & It was an important stop on Henry Louis Gates' "Miracle of the African World" special on \colorbox{red!20}{PBS}. \\
ChatGPT & This is an important stop on Henry Louis Gates' special program "The Wonders of the African World" on the \colorbox{red!20}{American Public Broadcasting System}. \\
GPT-4 & This is an important stop in Henry Louis Gates' \colorbox{red!20}{PBS} special program "Wonders of the African World." \\
\midrule
\textsc{Src} & \begin{CJK}{UTF8}{gkai} \colorbox{blue!20}{狼孩}如果完全由非人类的动物抚养长大，其行为（在身体条件允许的范围内）会与该动物非常雷同，比如会对人类表现出恐惧或冷漠。\end{CJK} \\
\textsc{Ref} & When completely brought up by non-human animals, the \colorbox{red!20}{feral child} exhibits behaviors (within physical limits) almost entirely like those of the particular care-animal, such as its fear of or indifference to humans. \\[0.5ex]
\hdashline
Google & If a \colorbox{red!20}{wolf child} is raised entirely by a non-human animal, its behavior (to the extent allowed by its physical condition) will be very similar to that animal, such as showing fear or indifference to humans. \\
ChatGPT & If a \colorbox{red!20}{wolf child} is raised completely by non-human animals, its behavior (within the limits of its physical conditions) will be very similar to that of the animal, such as showing fear or indifference towards humans. \\
GPT-4 & A \colorbox{red!20}{feral child} raised entirely by non-human animals would exhibit behavior (within the limits of their physical abilities) very similar to that of the animal, such as fear or indifference towards humans. \\
\bottomrule
\end{tabular}
\label{tab:analysis-case}
\end{table*}

\subsection{Case Study}

We present four test examples in Table~\ref{tab:analysis-case} for an intuitive understanding. The first example shows the hallucination of ChatGPT at the first few tokens and the inaccurate translation of  \begin{CJK}{UTF8}{gkai}过量降水\end{CJK}. 
The second example shows that both ChatGPT and GPT-4 translate \begin{CJK}{UTF8}{gkai}广泛耐药结核病\end{CJK} into the full name while the reference and Google Translate do not. 
The third example shows that GPT-4 can also translate the terminology \begin{CJK}{UTF8}{gkai}美国公共广播公司\end{CJK} into the abbreviation. 
The last example suggests that GPT-4 is able to translate the terminology \begin{CJK}{UTF8}{gkai}狼孩\end{CJK} more properly based on the context while Google Translate and ChatGPT fail to.

\section{Conclusion}
This work presents a preliminary study of ChatGPT for machine translation.
We find that ChatGPT performs competitively with commercial translation products (e.g., Google Translate) on high-resource European languages but lags behind significantly on low-resource or distant languages.
It also exhibits good results on spoken language while still performs worse than commercial systems on biomedical abstracts or Reddit comments.
We further explore an interesting strategy named \textbf{pivot prompting} that can improve the translation performance of distant languages noticeably.
With the launch of the GPT-4 engine, the translation performance of ChatGPT is significantly boosted, becoming comparable to commercial translation products, even for distant languages.
Extensive human analysis suggests that, ChatGPT has already become a good translator with GPT-4 as the Engine.

\section*{Limitations}

As a preliminary study, this work is far from complete with various aspects to make it more reliable:
\begin{itemize}
    \item \textbf{Comprehensiveness}: Currently, we randomly select 50 samples from each test set for evaluation due to the response delay of ChatGPT, which is not comprehensive due to the data coverage.
    Besides, we found that the results of the same query may vary across multiple trials, bringing randomness to the evaluation results. For more reliable results, it is best to repeat the translation multiple times for each test set and report the average result.
    \item \textbf{Translation Abilities}: We only focus on multilingual translation and translation robustness in this report. However, there are some other translation abilities that can be further evaluated, e.g., constrained machine translation and document-level machine translation.
\end{itemize}



\bibliography{anthology,custom}
\bibliographystyle{acl_natbib}



\end{document}